\documentclass{article}%
\usepackage{amsmath}
\usepackage{amsfonts}
\usepackage{amssymb}
\usepackage{graphicx}%
\providecommand{\U}[1]{\protect\rule{.1in}{.1in}}
\newtheorem{theorem}{Theorem}

\newtheorem{corollary}[theorem]{Corollary}

\newenvironment{proof}[1][Proof]{\noindent\textbf{#1.} }{\ \rule{0.5em}{0.5em}}
\begin{document}

\title{On Coarse Graining of Information and Its Application to Pattern
Recognition\thanks{Presented at MaxEnt 2014, 34th International Workshop on
Bayesian Inference and Maximum Entropy Methods in Science and Engineering
(September 21-26, 2014, Amboise, France).}}
\author{Ali Ghaderi\thanks{E-mail: Ali.Ghaderi@hit.no}\\\emph{{\small Telemark University College, Kj\o lnes Ring 56, NO-3918
Porsgrunn, Norway}}}
\maketitle

\begin{abstract}
We propose a method based on finite mixture models for classifying a set of
observations into number of different categories. In order to demonstrate the
method, we show how the component densities for the mixture model can be
derived by using the maximum entropy method in conjunction with conservation
of Pythagorean means. Several examples of distributions belonging to the
Pythagorean family are derived. A discussion on estimation of model parameters
and the number of categories is also given.

\end{abstract}

\section{\bigskip Introduction}

One of the goals of any scientific study is to identify regularities in
observations and classify them into possibly separate and simpler structures
or categories. These categories can in turn be used to make inferences on the
objects of interest. The major advantage of this approach is that one breaks
down a complicated reality into a collection of simpler structures. In a
similar way, in pattern recognition one is concern with discovery of
regularities in data but through use of computer algorithms which can be used
to classify the data into different categories \cite{BishopBook2006}.
Independent of ones point of view, any such analysis must start with
definition of the categories. If one has sufficient information about the
categories and their members, it is an easy task to establish a precise
definition. However, for most real life situations this is not the case and
the notion of category cannot be precisely defined. Under such conditions a
fruitful approach is to consider a category as collection of objects which are
likely to share the same properties. That is, in cases for which the
information available is insufficient to reach certainty, we ought to quantify
the degree to which we believe an object belongs to a given category. This
degree of belief is described by probability distribution over the space of
objects of interest, or sample space to be more precis.

The major bulk of the literature on the subject is dedicated to numerical
aspect of the problem. While acknowledging that the numerical challenges can
seriously compromise the applicability of a method, we believe that the
fundamental problem of modelling categories under partial knowledge condition
is just as important. In the following we will look at a class of problems in
pattern recognition for which one is in possession of empirical distribution
(histogram) over the objects of interests and a prior knowledge on the number
of categories involved. We propose an approach to modelling of the empirical
distributions based on the finite mixture models which relies on identifying
the relevant intensive properties of each category. In order to demonstrate
this method, we will show how conservation of Pythagorean means, the most
encountered class of intensive properties, in conjunction with maximum entropy
method can be used to derive the functional form of the mixture model. We will
also briefly discuss the extension to other conserved quantities and also give
a short overview on numerical challenges related to the inference problem. In
this article we restrict ourselves to positive univariate continuous quantities.

\section{Mixture model}

In the situations where categories cannot be defined precisely, the
probabilistic description might be the only possible option. In the
probabilistic framework, we can only talk about the likelihood of an object
belonging to a category. To this end, let us assume that by some experiment
the observation $X$ is made but it is not by itself sufficient to uniquely
determine which category it belongs to. For example, the observations can be
the height of people in certain region/country for which underlying categories
are the age groups that each individual might belong to. In such cases one
considers $X$ as a random variable and tries to model its probability density
function $p$. One approach to model $p$ is based on the so called \emph{finite
mixture models }\cite{TitteringtonBook1985}. The underlying assumption in this
approach is that $p$ is a \emph{convex combination }of $k$ densities in which
each density represents a single category. That is%
\begin{equation}
p\left(  \left.  x\right\vert \mathbf{\psi}\right)  =%
{\textstyle\sum\limits_{j=1}^{k}}
\pi_{j}\left(  \mathbf{\theta}_{j}\right)  f_{j}\left(  \left.  x\right\vert
\mathbf{\theta}_{j}\right)  ,\text{ }x\in\mathcal{X}
\label{MixtureDistributionDef}%
\end{equation}
where%
\begin{equation}%
{\textstyle\sum\limits_{j=1}^{k}}
\pi_{j}\left(  \mathbf{\theta}_{j}\right)  =1,\text{ }\pi_{j}\geq0
\label{MixingWeightsDef}%
\end{equation}
and%
\begin{equation}
\int_{\mathcal{X}}f_{j}\left(  \left.  x\right\vert \mathbf{\theta}%
_{j}\right)  dx=1,\text{ }f_{j}\left(  \left.  x\right\vert \mathbf{\theta
}_{j}\right)  \geq0
\end{equation}
and%
\begin{equation}
\mathbf{\psi=}\left(  \mathbf{\pi},\mathbf{\theta}\right)  =\left(  \left\{
\pi_{1},\ldots,\pi_{k}\right\}  ,\left\{  \mathbf{\theta}_{1},\ldots
,\mathbf{\theta}_{k}\right\}  \right)  .
\end{equation}
In such cases, one says that $X$ has a finite mixture distribution and that
$p$ is a finite mixture density function. The parameters $\pi_{j}$ are called
\emph{mixing weights} and $f_{j}$ the \emph{component densities} of the
mixture. In the context of pattern recognition, $k$ is the number of
categories and $f_{j}$ is the density function describing the distribution of
the members of the category $j$. It should be emphasis that the component
densities do not necessarily belong to the same family of densities. Each
component density represents our best guess about the structure of its
respective category for which its existence is independent of the other categories.

In order to be able to adopt the mixture model to a specific problem, given
that a priori one knows the number of categories, requires that one tackles
two different problems. The first problem is to determine how to achieve a
quantitative description of state of partial knowledge, i.e. determining the
functional form of the component densities. The second problem is to determine
$\mathbf{\psi}$ based on the available evidence, i.e. the empirical density.

\section{Determination of component densities}

In general, objects in the same category are more similar to each other than
to those in other categories. This similarity invokes the notion that there
are properties at the coarser level which distinguishes the categories from
each other. In fact, if we consider a category as a
homogeneous\footnote{Homogeneous in the sense that there is continuity between
various members of the group.} group in which the members are recognizably
similar, then it is reasonable to assume that the properties that distinguish
it from other categories should be intrinsic and independent of the
\emph{coarse graining }within the category itself. This coarse graining
property is the key concept in finding the component distributions.

In general, coarse graining is achieved by first grouping the elements of the
category into blocks, each having the same volume. Then following a
predetermined rule, each block is replaced with a single element representing
the elements of that block. This procedure is iterated \emph{ad infinitum}. We
call a property that is invariant under coarse graining as \emph{intensive}.
In this context, a category can be characterized and distinguished from others
by its intensive properties. Identifying the relevant intensive properties are
often challenging. Usually a less challenging approach is to first determine
the so-called \emph{extensive }properties of the category. An extensive
property is a property that is additive under coarse graining. That is, under
coarse graining, the elements that replace the blocks at each step, also
inherit the sum of each of extensive properties of their respective block
elements. Moreover, at each coarse graining step, due to similarity and
homogeneity conditions which exist among the members of a category, the
extensive properties scale independent of the choice of specific block. This,
in general, results in greatly reducing the complexity of the analysis.
However, it is conceivable that one might discover many extensive properties
which might not be relevant to the classification problem at hand. In this
respect, the choice of relevant properties are often problem dependent.
Nevertheless, identifying and describing an extensive property means that one
is able to find a function, up to a scaling factor, which captures the
essential features of that property. It can be shown that the expectation of
such a function is invariant with respect to coarse graining and hence it is
intensive. For example, particle mass is an extensive property of a system
consisting of a collection of particles. Whilst, the expected mass of a
particle is intensive. In the following, we shall call the intensive
properties that are expressed in the form of expectations as the
\emph{conservation laws}\footnote{We adopt the view held by Steiner
\cite{Steiner1978} that laws of conservation are simply not causal laws. They
provide constraints on what is allowed to happen.}.

\subsection{Conservation of Pythagorean means}

Let $g\left(  x\right)  $ denote a function representing an extensive property
of a category. Up to a scaling factor, some of the most encountered forms of
$g$ are%
\begin{equation}
g\left(  x\right)  =x\text{, }g\left(  x\right)  =x^{-1}\text{, }g\left(
x\right)  =\ln x.
\end{equation}
The expected values of these functions constitute the so-called
\emph{Pythagorean means}. The Pythagorean means are the arithmetic, geometric
and harmonic mean. More precisely, for a positive univariate continuous
variable with density $f,$ the Pythagorean means are defined as
\begin{subequations}
\begin{align}%
\mu
&  =\int_{0}^{\infty}xf\left(  x\right)  dx\label{ArithM}\\
\ln\gamma &  =\int_{0}^{\infty}\ln xf\left(  x\right)  dx\label{GeoM}\\
\eta^{-1}  &  =\int_{0}^{\infty}x^{-1}f\left(  x\right)  dx \label{HarM}%
\end{align}
where $%
\mu
,$ $\gamma$ and $\eta$ are the arithmetic, geometric and harmonic means,
respectively. In this regard, one can talk about two categories being similar
with respect to some of the Pythagorean means.

\subsection{Maximum entropy}

Although the conserved quantities restrict the possible distribution of
elements in a category, nonetheless, still there might be up to infinitely
many distributions that satisfy the constraints. We are interested in the
distribution that conserves the quantities of the interest while allowing
maximum degree of freedom on the non-conserved quantities. It can be shown
that among all the distributions that fulfill the constraints, the most
uncommitted distribution is the one with largest relative entropy $S$%
\end{subequations}
\begin{equation}
S\left[  f,q\right]  =-\int f\left(  x\right)  \ln\frac{f\left(  x\right)
}{q\left(  x\right)  }dx \label{Entropy}%
\end{equation}
where $f$ is the unknown distribution and $q$, also known as the \emph{prior},
defines what we mean by the uniform distribution in the sample space
$\mathcal{X}$ \cite{CatichaMonograph2012,Sivia1996}. This method of finding a
distribution is known as \emph{maximum entropy} or in short \emph{MaxEnt}.

\begin{theorem}
Let all the three Pythagorean means be conserved. Then the MaxEnt distribution
is%
\begin{equation}
f\left(  x;\lambda_{1},\lambda_{2},\lambda_{3}\right)  =\frac{q\left(
x\right)  }{Z_{q}\left(  \lambda_{1},\lambda_{2},\lambda_{3}\right)
}x^{\lambda_{3}-1}\exp\left(  -\lambda_{1}x-\lambda_{2}x^{-1}\right)
\label{GeneralPythagoreanDistribution}%
\end{equation}
where%
\begin{equation}
Z_{q}\left(  \lambda_{1},\lambda_{2},\lambda_{3}\right)  =e^{\lambda_{0}}%
=\int_{\mathcal{X}}q\left(  x\right)  x^{\lambda_{3}-1}\exp\left(
-\lambda_{1}x-\lambda_{2}x^{-1}\right)  dx
\end{equation}
is the \emph{partition function}, which acts as normalization factor.

\begin{proof}
This is equivalent to finding the maximum of the Lagrangian $L$ with respect
to $f$%
\begin{align}
L\left[  f\right]   &  =-\int_{\mathcal{X}}f\left(  x\right)  \ln
\frac{f\left(  x\right)  }{q\left(  x\right)  }dx-\left(  \lambda
_{0}-1\right)  \left(  \int_{\mathcal{X}}f\left(  x\right)  dx-1\right)
\label{Lagrangian}\\
&  -\lambda_{1}\left(  \int_{\mathcal{X}}xf\left(  x\right)  dx-%
\mu
\right)  -\lambda_{2}\left(  \int_{\mathcal{X}}x^{-1}f\left(  x\right)
dx-\eta^{-1}\right) \nonumber\\
&  -\left(  1-\lambda_{3}\right)  \left(  \int_{\mathcal{X}}\ln xf\left(
x\right)  dx-\ln\gamma\right) \nonumber
\end{align}
where $\lambda_{0}\ldots\lambda_{3}$ are the four Lagrange multipliers
corresponding to the four constraints\footnote{Note that $\lambda_{0}-1$ and
$1-\lambda_{3}$ are used instead of $\lambda_{0}$ and $\lambda_{3}$ as a
matter of convenience.}. It can be shown that maximizing the functional $L$ is
equivalent to solving the corresponding \emph{Euler-Lagrange} equation of the
calculus of variations \cite{Arfken2001} which results in statement of the theorem.
\end{proof}
\end{theorem}

We shall say a distribution that share the same functional form as
(\ref{GeneralPythagoreanDistribution}) belongs to \emph{Pythagorean family of
distributions}. Note that $q$ can be even improper with non-compact support as
long as the distribution in (\ref{GeneralPythagoreanDistribution}) is
normalizable. If we know, up to a normalization constant, the functional form
of the prior $q$ and the values of the Pythagorean means then $f$ can be
uniquely determined. Moreover, note that if $q$ is very narrow then $f\approx
q,$ whilst if $q$ is very broad then its influence is negligible and can be
considered to be the uniform distribution. The following corollary is a direct
consequence of Eq. (\ref{GeneralPythagoreanDistribution}).

\begin{corollary}
Let $q$ be the improper uniform distribution on the positive real line. Then
for all $x\in\mathbb{R}^{+}$
\begin{equation}
f\left(  x;\alpha,\beta,\lambda\right)  =\frac{1}{2\alpha K_{\lambda}\left(
\beta\right)  }\left(  \frac{x}{\alpha}\right)  ^{\lambda-1}\exp\left\{
-\frac{\beta}{2}\left(  \frac{x}{\alpha}+\frac{\alpha}{x}\right)  \right\}
,\lambda\in\mathbb{R},\alpha>0,\beta>0\label{GIGdist}%
\end{equation}
where $K_{\lambda}$ is the modified Bessel function of the second kind and%
\begin{equation}
\lambda=\lambda_{3},\alpha=\sqrt{\frac{\lambda_{2}}{\lambda_{1}}},\beta
=2\sqrt{\lambda_{1}\lambda_{2}}.
\end{equation}

\end{corollary}

\begin{figure}[tbp] \centering
\begin{tabular}
[c]{c}%
{\includegraphics[
height=2.4509in,
width=3.9167in
]%
{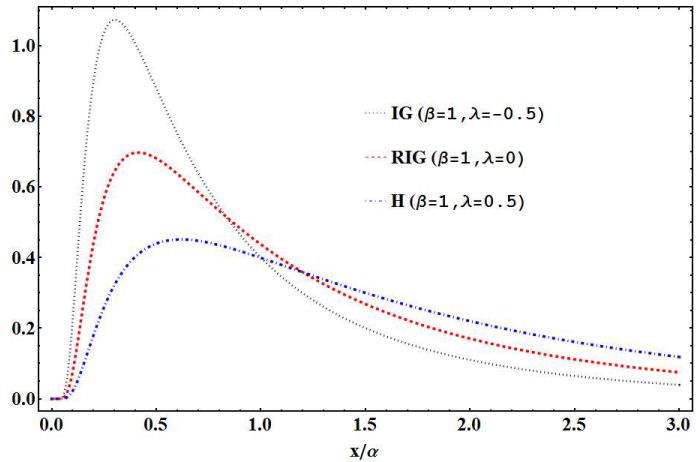}%
}
\end{tabular}
\caption{Three of the known  sub-classes of the generlized inverse Gaussian
distribution (GIG).}\label{FigureGIG}%
\end{figure}

In literature the distribution (\ref{GIGdist}) is known as \emph{generalized
inverse Gaussian (GIG) }distribution \cite{Jorgensen1982}. Some of its
well-known sub-classes are the \emph{inverse Gaussian (IG)} ($\lambda=-1/2$),
the \emph{reciprocal inverse Gaussian (RIG)} ($\lambda=-1/2$) and the
\emph{hyperbolic (H)} ($\lambda=0$ ) distributions (see Fig.\ref{FigureGIG}).%
Other familiar distributions arise when only some of the Pythagorean means are
conserved. For example, if one drops the constraint on arithmetic mean, that
is $\lambda_{1}=0$ in (\ref{Lagrangian}), the distribution is known as
\emph{inverse gamma }distribution. If the constraint on the harmonic mean is
dropped, that is $\lambda_{2}=0$ in (\ref{Lagrangian}), the distribution is
the \emph{gamma }distribution. The list is longer than this but the above
examples demonstrate the abundance of different variety of distributions
belonging to Pythagorean family.

\subsection{Other conservation laws}

For the sake of clarity we narrowed the discussions to the conservation of
Pythagorean means. But other conservation laws are possible and are even at
use. The MaxEnt method can handle other conserved quantities as well.
Nonetheless, it is recommended that one should always conduct an assessment on
conservation of Pythagorean means at the start of the analysis. The outcome
can be used as prior $q$ in (\ref{Entropy}) along with other conserved
quantities to derive the functional form of the component densities.

\section{Bayesian inference}

We have not touched the numerical aspect of this problem. It is often case
dependent and difficult to discuss without getting into the specifics.
However, the statement of the most important problems using the rules of
probability is quite simple.

\subsection{Determination of model parameters $\mathbf{\psi}$}

\begin{figure}[tbp] \centering
\begin{tabular}
[c]{c}%
{\includegraphics[
height=2.4933in,
width=3.9167in
]%
{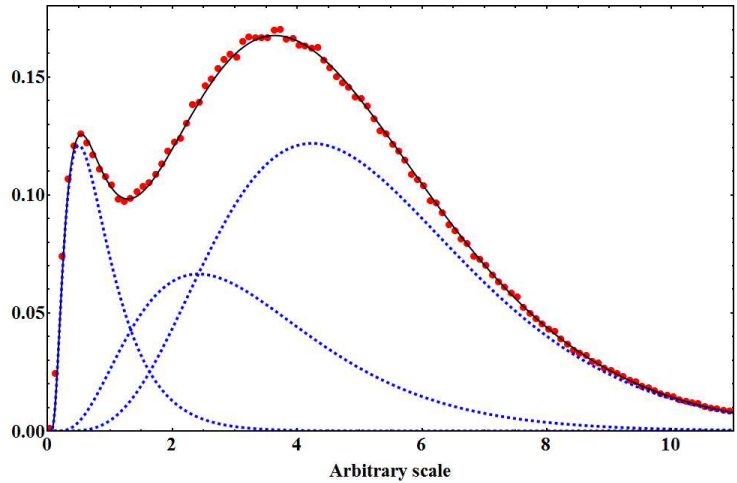}%
}
\end{tabular}
\caption{The dots represent the histogram of 50000 numbers, simulated from
the mixture of three GIG-variates. The dashed curves are the component
densities of the variates times their respective mixing weights. The sum of
the three dashed curves is the mixture model in red.}\label{FigureMixture}%
\end{figure}

Let $I$ summarize the information about the functional form of the component
densities and their number. Technically, once $I$ is known, determining
$\mathbf{\psi}$ in (\ref{MixtureDistributionDef}) becomes a standard problem
in statistical inference. To this end, assume that the observations are
randomly generated from $p\left(  \left.  x\right\vert \mathbf{\psi},I\right)
$. Then the normalized histogram of the data, say $h\left(  x\right)  $, can
be considered as the empirical estimate for $p\left(  \left.  x\right\vert
\mathbf{\psi},I\right)  $. Consequently, the unknown $\mathbf{\psi}$ can be
estimated from $h$ by using the Bayesian methods. Indeed, it follows from
\emph{Bayes rule} that%

\begin{equation}
p\left(  \left.  \mathbf{\psi}\right\vert h,I\right)  \propto p\left(  \left.
\mathbf{\psi}\right\vert I\right)  p\left(  \left.  h\right\vert
\mathbf{\psi,}I\right)  \label{PosteriorModel}%
\end{equation}
where $p\left(  \left.  \mathbf{\psi}\right\vert I\right)  $ is the prior for
$\mathbf{\psi}$. Usually we just have some rough knowledge about the domain of
$\mathbf{\psi}$ and therefore it is common to assume that $p\left(  \left.
\mathbf{\psi}\right\vert I\right)  $ is uniformly distributed over that
domain. The likelihood function $p\left(  \left.  h\right\vert \mathbf{\psi
,}I\right)  $ depends on our assessment of the sources that contribute to
deviation between the model and data and, in general, is problem specific
\cite{Gregory2005,Udo2014}. The most likely estimate for $\mathbf{\psi}$ is
the one which coincides with the global maximum of $p\left(  \left.
\mathbf{\psi}\right\vert h,I\right)  $ which is called \emph{maximum a
posteriori probability (MAP) estimate}. Usually, due to intractability of
analytical form of the posterior distribution the methods for estimating MAP
are \emph{Monte Carlo} based \cite{RobertCasellaBook2005}. For illustration
purpose, in Fig. \ref{FigureMixture}, we have plotted an example of a mixture
model and its three component densities versus their joint simulated
histogram.%

\subsection{Determination of number of categories $k$}

In the above discussions, we assumed that the number of categories are known.
However, often we do not know this number and we need to estimate it. In the
Bayesian framework this is known as \emph{model selection problem}. Indeed, in
order to estimate the number of categories we need to evaluate the posterior
distribution for $k$ conditional on $h$. By the Bayes rule we have%
\begin{equation}
p\left(  \left.  k\right\vert h,I^{\prime}\right)  \propto p\left(  \left.
k\right\vert I^{\prime}\right)  p\left(  \left.  h\right\vert k\mathbf{,}%
I^{\prime}\right)
\end{equation}
where $I^{\prime}$ summarize the information about the functional form of the
component densities. Note that $I=\left(  k,I^{\prime}\right)  $. Now, by
\emph{marginalization} and \emph{product rule }we have%
\begin{equation}
p\left(  \left.  h\right\vert k\mathbf{,}I^{\prime}\right)  =\int_{\Psi
}p\left(  \left.  h,\mathbf{\psi}\right\vert k\mathbf{,}I^{\prime}\right)
d\mathbf{\psi}=\int_{\Psi}p\left(  \left.  \mathbf{\psi}\right\vert I\right)
p\left(  \left.  h\right\vert \mathbf{\psi,}I\right)  d\mathbf{\psi}%
\end{equation}
and hence%
\begin{equation}
p\left(  \left.  k\right\vert h,I^{\prime}\right)  \propto p\left(  \left.
k\right\vert I^{\prime}\right)  \int_{\Psi}p\left(  \left.  \mathbf{\psi
}\right\vert I\right)  p\left(  \left.  h\right\vert \mathbf{\psi,}I\right)
d\mathbf{\psi}. \label{ModelSelection}%
\end{equation}
The integral on the right hand side of (\ref{ModelSelection}) is known as
\emph{evidence} and is equal to normalization factor on the right hand side of
(\ref{PosteriorModel}). If one assumes $p\left(  \left.  k\right\vert
I^{\prime}\right)  $ to be uniform then the most probable value of $k$ is the
one which corresponds to the model with largest evidence. It is often quite
challenging to get a good estimate of evidence. Most methods are Monte Carlo
based and have their own pros and cons. Therefore, the choice of the method is
very much application dependent. It is not uncommon that one uses several
different methods in order to find a good estimate. For an overview over the
most used methods the reader is referred to \cite{Friel2012}.

\section{Conclusion}

In situations where we have partial knowledge about the categories, the
probabilistic description based on the finite mixture model is a possible
approach. In order to determine the component densities of the model one can
start with finding the relevant extensive properties of each category under
coarse graining. Taking the expectation of these extensive properties will
lead to the right conservation laws and in conjunction with MaxEnt to
component densities. Then the model parameters can be estimated from the
empirical density data using the standard Bayesian methods.

\section{Acknowledgments}

I would like to thank Nicholas Armstrong for insightful advice on calculation
of evidence.

\bibliographystyle{alpha}
\bibliography{CoarsegrainingBlank}

\end{document}